\documentclass{article}

\usepackage{microtype}
\usepackage{graphicx}
\usepackage{booktabs}
\usepackage{amsfonts}       
\usepackage{nicefrac} 
\usepackage{amsmath}
\usepackage{amssymb}
\newcommand{\R}{\mathbb{R}}
\usepackage{subcaption}
\usepackage[normalem]{ulem}
\useunder{\uline}{\ul}{}
\usepackage{algorithm}
\usepackage{algorithmic}
\usepackage{adjustbox}
\usepackage{array}
\usepackage{color}
\usepackage{amsmath}
\usepackage{IEEEtrantools}
\usepackage{mathtools}
\usepackage{multicol}
\usepackage{siunitx}
\usepackage{hyperref}

\usepackage[accepted]{icml2020}

\icmltitlerunning{Stealing Black-Box Functionality Using The Deep Neural Tree Architecture}

\begin{document}

\twocolumn[
\icmltitle{Stealing Black-Box Functionality Using The Deep Neural Tree Architecture}

\icmlsetsymbol{equal}{*}

\begin{icmlauthorlist}
\icmlauthor{Daniel Teitelman*}{Rafael}
\icmlauthor{Itay Naeh*}{Rafael}
\icmlauthor{Shie Mannor}{Technion}
\end{icmlauthorlist}

\icmlaffiliation{Rafael}{Rafael - Advanced Defense Systems Ltd., Israel}
\icmlaffiliation{Technion}{Department of Electrical Engineering, Technion, Israel}

\icmlcorrespondingauthor{Daniel Teitelman}{danieltay@rafael.co.il}
\icmlcorrespondingauthor{Itay Naeh}{itay@naeh.us}
\icmlcorrespondingauthor{Shie Mannor}{shie@technion.ac.il}

\icmlkeywords{Machine Learning, Deep Learning , ICML, Black-box, Black Box, Neural Arithmetic Logic Units, Regression Tree, Decision Tree, Cloning, Active Learning, Explainable AI, Stealing Functionality}

\vskip 0.3in
]

\printAffiliationsAndNotice{\icmlEqualContribution} 

\frenchspacing

\begin{abstract}
This paper makes a substantial step towards cloning the functionality of black-box models by introducing a Machine learning (ML) architecture named Deep Neural Trees (DNTs). This new architecture  can learn to separate different tasks of the black-box model, and  clone its task-specific behavior. We propose to train the DNT using an active learning algorithm to obtain faster and more sample-efficient training. In contrast to prior work, we study a complex "victim" black-box model based solely on input-output interactions, while at the same time the attacker and the victim model may have completely different internal architectures. The attacker is a ML based algorithm whereas the victim is a generally unknown module, such as a multi-purpose digital chip, complex analog circuit, mechanical system, software logic or a hybrid of these. The trained DNT module not only can function as the attacked module, but also provides some level of explainability to the cloned model due to the tree-like nature of the proposed architecture.
\end{abstract}

\raggedbottom

\section{Introduction} \label{Introduction}

The black-box cloning problem as defined by \cite{Knockoff} is the problem of reconstructing or ``stealing" the functionality of a system using only black-box interactions, which means we only observe the input-output pairs. In the paper above, the problem of black-box cloning is only discussed for a victim which is based on deep learning (DL), where the attacker (adversary) also has the same general type of architecture. In contrast to this prior work, our work will focus on a more general case of the problem where the attacker is employing a ML-based model and the victim is an arbitrary module which preforms different tasks. We would like to clone the complete black-box functionality with our ML-based model, by using hardware based black-box model as a test case.
Within this formulation, we  answer the following questions in our paper with an end goal of cloning black-box models:
Is it possible to train a clone on a random set of query inputs corresponding to the black-box predictions? Could we differentiate between the different tasks preformed by the black-box model? And is it possible to learn a specific task preformed by the black-box model after differentiating it from other tasks? Another question that arises is whether it is possible to train a clone in an efficient way: Is it possible improve sample efficiency of queries  without any human in the training loop?

\begin{figure}[ht!]
  \includegraphics[width=\columnwidth]{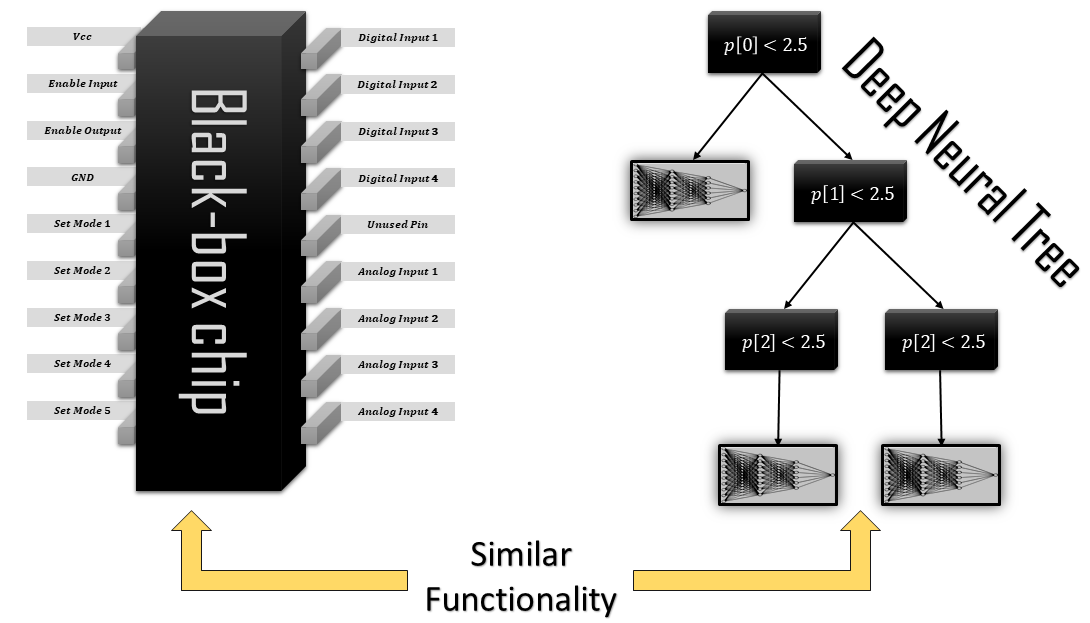}
  \caption{An attacker can create a clone of a black-box model solely by interacting with it by input output interactions.}
  \label{fig:BlackBoxChipDNTinteraction}
\end{figure}

Moreover, another interesting question is which ML architectures provide a good basis for a cloned model? In particular, this calls for {\em explainable AI}: Is it possible to create an architecture that also explains how the black-box it cloned works?
With this questions in mind, we will present in this paper an architecture the Deep Neural Tree (DNT) that is able to clone black-box models trained by an active learning algorithm, while also having the ability to explain how  the black-box we cloned may work. The DNT architecture will not only be able to clone general black-box models with hierarchy but also solve regression problems with intrinsic hierarchy. Our contributions include presenting a novel architecture that generalizes three well-established architectures and combines the  benefits of each one, while incorporating training by an active learning algorithm. The DNT architecture allows us to clone chips that at a speed and accuracy beyond the state-of-the-art.

\section{Background} \label{Background}

In this paper we use a combination of three different architectures: decision trees, neural networks, and Neural Arithmetic Logic Units (NALU) \cite{NALU}. 
The idea behind combining three different architectures so different from each other in a single new architecture, arises as a direct result of the advantages each architecture holds. Decision trees can be understood as function discriminators which will allow us to distinguish between the different tasks the black-box preforms. Neural networks allow us to learn complex models we have no knowledge about, while the NALU module will allow our neural networks to learn more complicated functions and at the same time extrapolate better. In order to demonstrate the DNT architecture's ability to clone black-box models we will use a digital multi purpose chip, which will be described in section \ref{Deep Neural Tree}. In addition, we will discuss in section \ref{Experiments} active learning and explainable AI as they were important factors to keep in mind while developing the DNT architecture.

\subsection{Regression Trees}
Decision trees are a non-parametric supervised learning method used for classification and regression \cite{Trees}. The goal is to create a model that predicts the value of a target variable by learning simple decision rules inferred from the data features. Training decision trees is done as follows: Given training vectors \begin{math} x_i \in \R^n , i=1,...,l \end{math} and a label vector \begin{math} y \in \R^l \end{math} a decision tree recursively partitions the spaces such that the samples with the same label are grouped together. Let the data at node $m$ be represented by $Q$. For each candidate split \begin{math} \theta=(j,t_m) \end{math} consisting of a feature $j$ and a threshold $t_m$, partition the data into \begin{math} Q_{left}(\theta) \end{math} and \begin{math} Q_{right}(\theta) \end{math} subsets:

\vskip -0.1in
 
\small
\begin{equation*}
    Q_{left}(\theta)=(x,y)|x_j \leq t_m  ,  Q_{right}(\theta)=Q \backslash Q_{left}(\theta)
\end{equation*}
\normalsize

The impurity at $m$ is computed using an impurity function $H()$, which in our case is defined as followed:

\vskip -0.1in
\begin{equation*}
    H(X_m)=\frac{\sum_{i \in N_m} (y_i-\bar{y}_m)^2}{N_m} , \bar{y}_{m}=\frac{\sum_{i\in N_m} y_i}{N_m}
\end{equation*}

This impurity function represents our regression criteria, for a continues value in node $m$, which represents a region $R_m$ with $N_m$ observations, under the common criteria to minimise the $L_2$ error (MSE). The impurity function definition allows us to define the following cost function:

\vskip -0.1in
\small
\begin{equation*}
    G(Q,\theta)=\frac{n_{left}}{N_m}H(Q_{left}(\theta))+\frac{n_{right}}{N_m}H(Q_{right}(\theta))
\end{equation*}
\normalsize

We select the parameters that minimises the cost function:

\vskip -0.1in
\begin{equation*}
    \theta^*=  \arg\min_{\theta}G(Q,\theta)
\end{equation*}

The training process continues by recursion over the subsets $Q_{left}(\theta^*)$ and $Q_{right}(\theta^*)$ until the maximum allowable depth is reached or $N_m=1$.

We use decision trees as this function discriminator. This means that we use a decision tree to learn the different modes the chip has, while not learning their exact input-output behavior as it will make the decision tree overfit the training data. An important thing to note is the choice of a regression tree over a classification tree that was done as direct result of our model being {\em both} continuous and discrete. A fact which makes a regression tree better suited towards this kind of task; see more discussion below. 

\subsection{Neural Arithmetic Logic Units}

While neural networks can successfully represent and manipulate numerical quantities given an appropriate learning signal, the behavior that they learn may not generalize well in some cases. Specifically, one frequently observes failures when quantities lie outside the numerical range used during training while testing.  Neural Arithmetic Logic Units (NALUs)  were proposed by \cite{NALU}, as a new module that could be add added to any neural network architecture. NALUs help in generalizing quantities to neural network and help the models generalize for tasks like extrapolation. The NALU model consists of another module, the Neural Accumulator (NAC), implemented as follows:

\vskip -0.2in
\begin{align*}
    \text{NAC}: \quad& \textbf{a} = \textbf{W}x \\
                &\textbf{W} = tanh(\hat{\textbf{W}})\odot \sigma ( \hat{\textbf{M}}) \\
    \text{NALU}:\quad& \textbf{g}\odot \textbf{a} + (1-\textbf{g})\odot \textbf{m} \\
                &\textbf{m} = exp \; \textbf{W}(log(|\textbf{x}|+\epsilon)), \textbf{g} = \sigma (\textbf{Gx})
\end{align*}

Where $\sigma(\cdot)$ is the sigmoid function $\hat{\textbf{W}},\hat{\textbf{M}},{\textbf{G}}$ are tensors and $\epsilon$ prevents log 0.
In our architecture we use the NALU as a layer inside our neural network, this module as described above will help us extrapolate better in addition, it will allow us to learn more complicated functions as the NALU module will allow the neural network to have a higher capacity for learning functions, in comparison to traditional neural network. 

\begin{table*}[!ht]
\centering
\begin{tabular}{|c|c|c|c|c|}
\hline
{\ul \textbf{PIN \#}} & {\ul \textbf{Pin Name}} & {\ul \textbf{Type}} & {\ul \textbf{Values}} & {\ul \textbf{Description}}                     \\ \hline
p{[}0{]}              & Vcc                     & Power               & {[}0,5{]} V           & From GND                                       \\ \hline
p{[}1{]}              & Enable Input            & Binary              & {[}0,5{]} V           & enb 1,2                                        \\ \hline
p{[}2{]}              & Enable Output           & Binary              & {[}0,5{]} V           &                                                \\ \hline
p{[}3{]}              & GND                     &                     & {[}0,5{]} V           & From Vcc                                       \\ \hline
p{[}4{]}              & Set Mode 1              & Binary              & {[}0,5{]} V           & Amplify input 9 by input 10 p{[}9{]}*p{[}10{]} \\ \hline
p{[}5{]}              & Set Mode 2              & Binary              & {[}0,5{]} V           & Read from LUT1 position of input 11            \\ \hline
p{[}6{]}              & Set Mode 3              & Binary              & {[}0,5{]} V           & Read from LUT2 position of input 12            \\ \hline
p{[}7{]}              & Set Mode 4              & Binary              & {[}0,5{]} V           & Average(LUT1(input 11),LUT2(input 12))         \\ \hline
p{[}8{]}              & Set Mode 5              & Binary              & {[}0,5{]} V           & Binary from inputs 14, 15, 16, 17              \\ \hline
p{[}9:12{]}           & Inputs                  & Analog              & {[}0,10{]} V          &                                                \\ \hline
p{[}13{]}             & Unused Pin              & -                   &                       & Doing nothing                                  \\ \hline
p{[}14:17{]}          & Inputs                  & Binary              & {[}0,5{]} V           & Inputs                                         \\ \hline
\end{tabular}
\caption{Multi purpose digital chip - the black-box we will clone\label{t:chip}} 
\end{table*}

\subsection{The Chip}

We simulated the following electronic chip\footnote{\href{https://github.com/danielt17/Deep-Neural-Trees-Simulation-Environment/tree/4499294ddcfd77a4777951c9c5c49b340a54b2dc}{Digital chip simulation environment}} shown in Table \ref{t:chip} which we will use as our black-box victim. The chip's specifications for safe voltage use, and function definitions are shown in Equations \ref{eq:1} to \ref{eq:4}.

\begin{align}
    \shortintertext{Binary Threshold:} 
    -1.0:2.5 = False, 2.5:6.0 = True\label{eq:1}\\
    \shortintertext{Vcc-GND:}
    -1.0:4.0 = False, 4.0:6.0 = True\label{eq:2}\\
    \shortintertext{Lookup Table 1:}
    4\cdot in^2 + 0.1\cdot in^3 + 0.1\cdot in^4 + 7 \quad\label{eq:3}\\
    \shortintertext{Lookup Table 2:}
    14\cdot \sin(in)\cdot e^{-2\cdot in} \quad \quad \quad \quad \quad \quad \: \: \:\label{eq:4} 
\end{align}

\vskip -0.1in
The  difficulties with cloning a multi-purpose digital chip arise from inherently different problems. We list some of this problems and point to them for the chip from Table \ref{t:chip}.
First, the chip has both analog and digital inputs.  The digital inputs typically define the different modes of the chip and the analog ones define a continues scale of outputs for the chip. 
Second, the chip has many different modes with a hierarchy between them where the specified mode is defined as the most significant bit of the vector $p[4:8]$; this definition with a combination of the other enable bits, make this model hard to learn for traditional neural networks. 
Third, the training data will be skewed as most inputs will be discarded for learning because the input and output enables mask the actual function. This makes ``useful" data a lot scarcer for learning thus, requiring an active learning training algorithm as described in \cite{ActiveLearningSurvey}. The three problems described above limit the effectiveness of traditional neural networks for solving the black-box problem.
This leads us to introduce in section \ref{Deep Neural Tree} the DNT architecture that have the ability to overcome the three difficulties mentioned above and solve black-box cloning problems.

\subsection{Active Learning}

As discussed by \cite{ActiveLearningSurvey} active learning is the ability of system to be "curious" in regards to the training data it selects for training, this idea arises from the hypothesis that if an algorithm selects its own training data than it will train faster while also disregarding "useless" training examples. In our DNT architecture as discussed in section \ref{Deep Neural Tree} we introduced a new active learning algorithm in order to select meaningful training data while also shortening the training time. 

\section{Related Work} \label{Related Work}

\textbf{Model cloning:} Model cloning or model  stealing using black-box ML-based models has  recently gained popularity: parameters \cite{SecurityPrivacy}, hyper-parameters \cite{Hyperparamters}, architecture \cite{ReverseBB}, information on training data \cite{Membership} and decision boundaries \cite{PracticalBB} and even whole ML models \cite{Knockoff,gradientBB}. In contrast to the methods discussed above that investigate the possibility of cloning or stealing black-box ML models using ML, we propose a new architecture which solves the black-box problem in a more general case for models that have no inherit connection to the neural network architecture. This problem is inherently more difficult because the attacker and the victim have two different architectures: The attacker is a neural network and the victim in this case is a ``simple" chip with many different operations with almost no connection between them. Furthermore, we assume we have no knowledge regarding the distribution of inputs given to the chip, which leads to a skewed data set in task-space, forcing us to find an efficient way to train the architecture. This idea is discussed in a paper by \cite{bbActiveLearning} that focuses on active learning in order to attack black-box function interfaces efficiently to hijack the black-box behavior. In our paper we will perform active learning in the DNT architecture by creating an algorithm based on decision trees, neural networks and the NALU module by selecting the ``correct" inputs for fast and sample efficient learning. The idea of active learning in combination with decision trees was also suggested in \cite{ActiveLearningTrees}, however we use a different method based upon the classical DFS algorithm.

Another issue our architecture attempts to addresses is that of explainability. We would like to learn the input-space combinations that switch between the different tasks the chip is programmed to perform and what decision or input path we should take in order to map them. We achieve explainability in this regard by using decision trees in combination with neural networks an idea discussed recently by \cite{AdaptiveNeuralTree} which suggested a model combining neural network with decision trees in order to learn hierarchical structures while maintaining explainable nature of the resulting architecture. This idea was presented even earlier by \cite{NeuralTreeClass}. There are similarities in the core ideas behind them as we all want to combine the power of decision trees with neural networks but, their work is inherently different as we do not combine the neural network with a decision tree, and instead we use the decision tree for task differentiation and training, essentially as a pre-processor for our neural networks. Finally the most important problem we  solve is that of creating an architecture, able to learn efficiently by an active learning training algorithm inherently different functionalities while also being explainable thus, solving all of the above an idea which have not been discussed in the above papers.

\section{Problem Statement} \label{Problem Statement}

\begin{figure}[ht!]
  \includegraphics[width=\columnwidth]{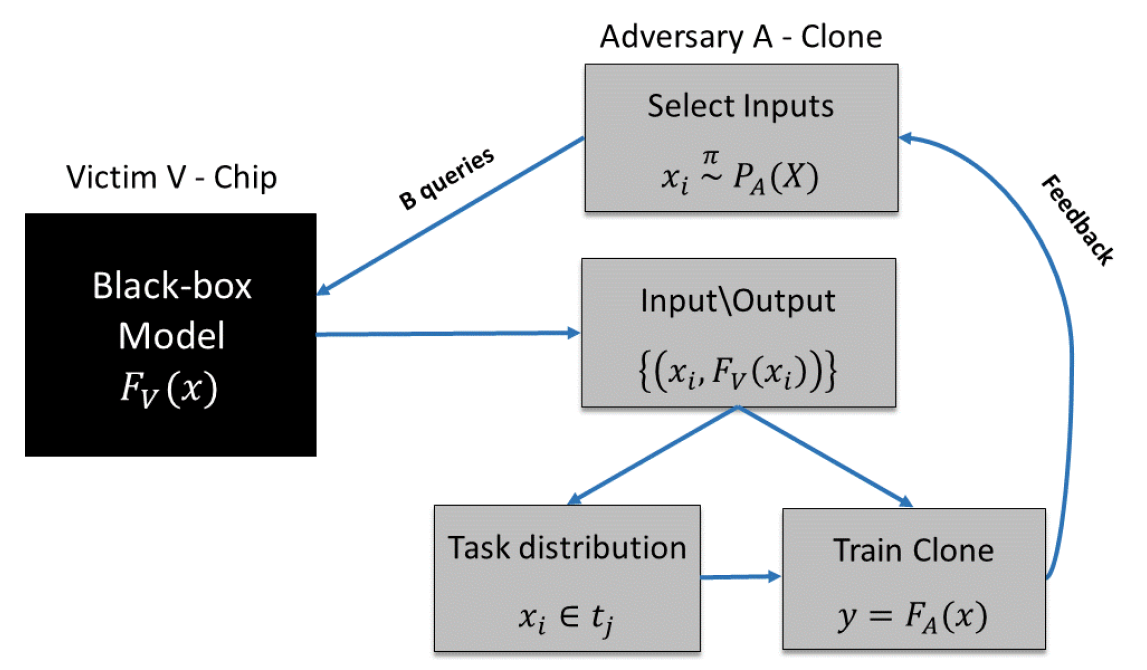}
  \caption{Laying out the task of model cloning in the view of two players: victim $V$, the chip, and an adversary $A$. We use training data generated by $V$ and train with it our clone $A$.}
  \label{fig:BlackBoxGameDefinition}
\end{figure}

We now formalize the task of functionality stealing as defined by \citet{Knockoff} while generalizing it to all types of black-boxes (ML based or not).
Given black-box query access to a ``victim" model $F_V: \mathcal{X} \rightarrow \mathcal{Y}$, to clone its functionality using ``clone" model $F_A$ of the adversary. As shown in Figure \ref{fig:BlackBoxGameDefinition}, we set it up as a two-player game between a victim $V$ and an adversary $A$. Now, we discuss the assumptions in which the players operate and their corresponding moves in this game.

\textbf{Victim's Move:} the victim's end goal is to perform some given mathematical operation on input $x_i$ and output the result $F_V (x_i)$.

\textbf{Adversary's Unknowns:} the adversary is presented with the black-box model that given any (high dimensional) input, outputs a one dimensional output that corresponds to the given operation. The only thing we assume to know is which voltages cause the chip to malfunction, therefore we will use them as a guideline for our model. This is a reasonable assumption with the game we setup due to the fact that an expert could tell us the boundaries for safe input voltages.

\textbf{Adversary's Attack:} to train our clone, the adversary: (i) randomly selects input $x_i \overset{\pi}{\sim} P_A (X)$ using strategy $\pi$ to obtain a training set of inputs and outputs: ${(x_i,F_V (x_i))}_{i=1} ^B$ where $B$ is the set size. (ii) The architecture splits the training set in relation to the different tasks, each input produces $x_i \in t_j$ where $t_j$ is some task $t$ numbered $j$. (iii) After splitting the training set to different tasks we train each sub-architecture according to the given task. (iv) Gains feedback for further more optimised training.

\textbf{Objective:} our main focus is the adversary, whose primary objective is training a clone that preforms well on the tasks the chip is designed for thus, cloning it. We have also three secondary objectives: (i) {\bf Efficient sampling.} Some black-boxes have a physical limit on their usage. This limit is extremely strict as in real world uses, you would only be able to sample a small amount of times the black-box before it malfunctions. (ii) {\bf Efficient task separation.} The black-box model we investigate has its operating modes cascaded, thus requiring a novel architecture to solve this separation task. (iii) {\bf Explainable AI.} We would like to understand the internal logic of the given black-box using our clone, as our human expert would like to have some insight into the black-box architecture the DNT architecture clones. Our main objective follows as direct consequence of our secondary objectives, as we want to create an architecture to solve this task.

\section{Deep Neural Tree} \label{Deep Neural Tree}

\subsection{The Model}

Here we propose a deep learning architecture with the ability to clone black-boxes without memory (the output of the black-box does not depend on previous inputs). Our architecture has the following abilities: to distinguish between different operations done by the black-box, learning by efficient sampling and getting better at sampling the longer the probing process is advancing by an active learning algorithm. Extrapolate the learned functions for values outside the scanned parameter space and the ability to explain the user the distilled intrinsic logic of the black-box.

\begin{figure}[ht!]
  \includegraphics[width=\columnwidth]{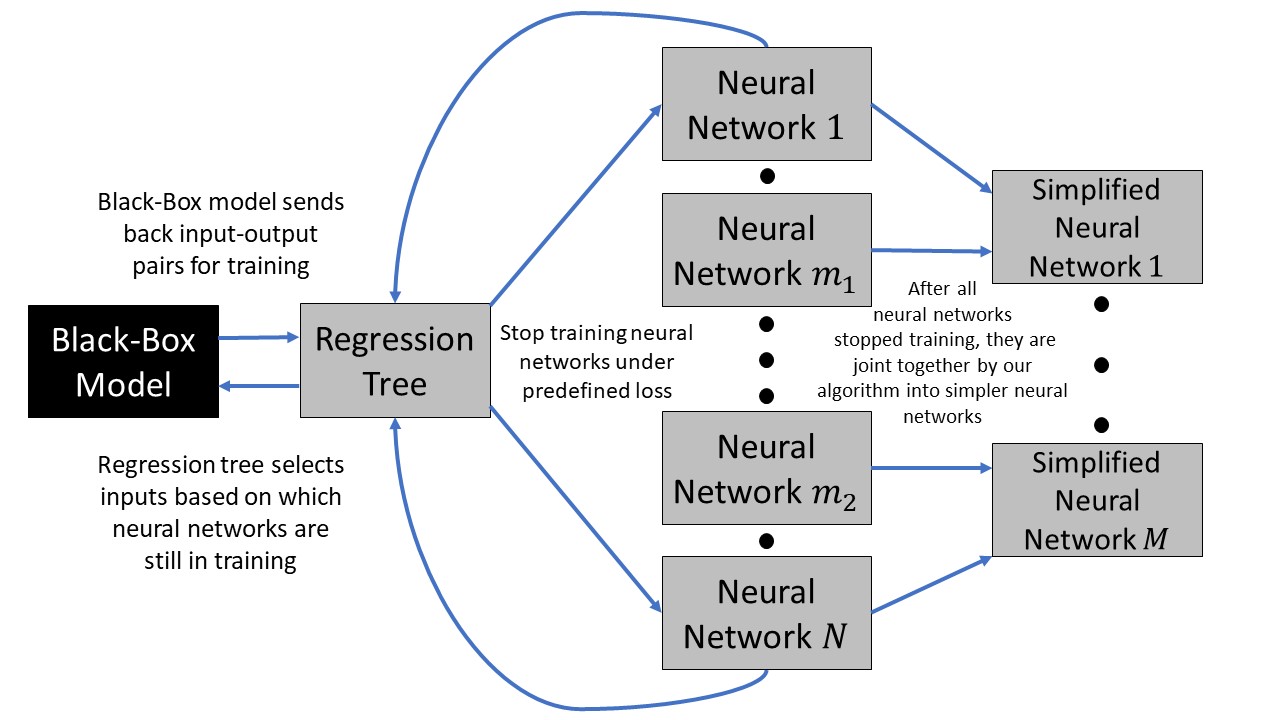}
  \caption{An illustration of the training scheme for the Deep Neural Tree architecture.}
  \label{fig:DNTtrainingProcedure}
\end{figure}

Our architecture combines the ability of the regression tree to distinguish between different tasks and the neural network ability to learn an unknown model. We train our regression tree with respect to the training set produced by querying the black-box. Then we create a NALU embedded neural networks, which will provide higher learning capacity at each leaf of the tree. Typical neural network will consist of 3 fully connected RELU activated layers, a NALU module and two more fully connected RELU activated layers. Each neural network will be tasked with a smaller task defined by the corresponding leaf. The reasoning behind this architecture is splitting the general high-level tasks by the branches of the tree and performing the low-level tasks using the deep neural networks.

\subsection{Training Methodology}

Using this architecture we can start training the module. We use our trained regression tree for splitting the training set into smaller training sets, where each training set is given to the neural network embedded in the corresponding leaf. For reducing training time and minimizing the black-box queries we use a Depth-First-Search (DFS) algorithm on the regression tree in order to map the logical path from each leaf back to the input space, and more important, from the input-space to each leaf. This will allow us to provide more insightful inputs that will reach the leaves with the networks that demonstrate higher loss values. Networks with sufficiently low loss levels will stop the training and stop querying input data. 
This will assists the convergence of the learning process by asking the most relevant question in input-space and training only networks that still needs to lower their loss values. We continue with this training algorithm until all of the neural networks reached the desired loss values or maximum amount of iterations.

\begin{algorithm}[!ht]
   \caption{DNT Architecture Training Procedure}
   \label{alg:DNTtraining}
\begin{algorithmic}[1]
    \small
    \STATE Initialize $error$, and $m$ as number of input pins.
    \STATE Train regression tree of depth $n \leq m$ with randomly generated data from uniform distribution$~U[V_{min},V_{max}]^{m}$.
    \STATE Define $l$ as the amount of leaves in the regression tree.
    \STATE Create a list of $l$ neural networks named $netList$.
    \STATE Define array named $flaggedTrainingSet$ with $m+1$ columns.
    \REPEAT
        \STATE Generate data from$~U[V_{min},V_{max}]^{m}$ and add it to the $flaggedTrainingSet$ in the first $m$ columns.
        \STATE Pass data through regression tree, mark the data to its corresponding leaf - neural network location.
        \FOR{i in netList \& still in training}
            \IF{$loss(netList(i)) < error$}
                \STATE For each input sample in $flaggedTrainingSet$ add to the $m+1$ column the number $i$. 
            \ELSE
                \STATE Train $netList(i)$ with new training data corresponding to $i$-th neural network.
                \STATE Add to each training sample used in training the $i$ neural network, a flag of $i$ at the $m+1$ column in $flaggedTrainingSet$ array.
            \ENDIF
        \ENDFOR
    \UNTIL{Maximum number of iterations or satisfactory loss is reached}
\end{algorithmic}
\end{algorithm}

\subsection{Simplifying The Architecture}

The DNT architecture should start training with some redundancy in order to contain the complexity of the unknown black-box's functionality. In later stages of the training, when the general functionality was mapped and partially trained, we can reduce this redundancy by combining neural networks which posses the same functionality over similar input parameters. In order to do so we use a given input and evaluate the loss of all the networks. Then we effectively take a sub-group of neural networks with the most frequent occurrence and reduce them into a single network. We train the combined network on previously queried data that was used to train the separated networks, thus reducing additional queries of the black-box. This will allow not only the reduction of the number of trainable networks, but achieve a better generalization over larger input space. See algorithm \ref{alg:DNTsimplify} for further explanation.

\subsection{DNT \& Explainable AI}
The DNT architecture uses a combination of regression tree and neural networks, thus we get the advantages of both neural network to learn a model in a parametric free way and the ability of the regression tree to distinguish between different tasks while also being explainable to the user \cite{SurveyExplainableAI} by using DFS-scan algorithm that will allow the mapping of the decision process done by our regression tree.

\begin{algorithm}[!ht]
   \caption{DNT Architecture Simplifying Procedure}
   \label{alg:DNTsimplify}
\begin{algorithmic}[1]    
    \small
    \STATE Initialize search depth $n_1<n$.
    \STATE Define $l_1$ as the amount of nodes in the regression tree of depth $n_1$.
    \STATE Initialize an array of size $l_1$ named $rankTraining$ which defines the limit of how many neural networks can train a given node. 
    \STATE Create a list named $netSimplifiedList$ of $l_1$ neural networks.
    \STATE Run DFS algorithm to get a mapping of the decisions taken by the regression tree.
    \FOR{$j < l_1$}
        \STATE Using the DFS mapping to and from node $j$ of depth $n_1$ we will define the decision process which leads to node $j$ as $jAncestor$, and the neural networks which are the childs of node $j$ as $jChild$.
        \STATE We will search in $flaggedTrainingSet$ for all training examples with the $jAncestor$ path while adding them to an array named $jNodeTraining$.
        \FOR{every training example $p$ $ \in$ $jNodeTraining$}
        \STATE Search for the best neural network in $jChild$ under the following criteria: $argmin_{i}||y_{predict}^{i}(p)-y_{target}(p)||$ where $||.||$ is the $l_2$ norm.
        \STATE We will change the $m+1$ column in $jNodeTraining$ for our $p$ training example to hold the number of the neural network with lowest error.
        \ENDFOR
        \STATE Define array named $jNodeSimplifiedTraining$ which contains training samples ranked up to $rankTraining(j)$ by occurrence of the neural network number in the $m+1$ column of the $jNodeTraining$ array.
        \STATE We train now the $netSimplifiedList(j)$ by training data from $jNodeSimplifiedTraining$
    \ENDFOR
    \STATE Switch previous netList with the netSimplifiedList, connecting to its its analog $n_1$ depth node.
\end{algorithmic}
\end{algorithm}

\section{Experiments} \label{Experiments}

The experiments in this section were implemented in TensorFlow\footnote{\url{https://www.tensorflow.org/}} and Scikit-Learn\footnote{\url{https://scikit-learn.org/}} while running on an intel i7-6700 CPU. We tested the ability of the DNT architecture to clone the functionality of the chip accurately, while minimizing learning time and queries.
Several tasks are required: distinguishing between different operations (operation modes), learning a specific task accurately and improving its sampling efficiency. Regular deep neural networks are very good at learning and generalizing various specific tasks. Alas, training a DNN on several different tasks proved to be problematic. 
This is why we Incorporated regression tree within the DNT, where a set of smaller, task-specific networks are operated by the logic of the regression tree, which is preferable in selective hierarchical task conditioning.

\textbf{Mathematical complexity:} Our architecture has the ability to clone entire black-box models, but in this section we will demonstrate this ability on two particular selects (operation modes): select \#2 and select \#3 of the black-box chip. These two selects are easy to learn individually, but they are very different when learned together on a single network. Beside learning the individual modes, the DNT needs to learn the 1-hot coded and cascaded mode selection of the chip.\\
\textbf{Skewed data set:} Due to the cascaded nature of the 1-hot mode selection, the randomly generated inputs "waste" 75\% of the data set on enabling the "enable input" pin and the "enable output" pin. From the remaining 25\% contributing examples, 50\% goes into selection \#1, 25\% into selection \#2, 12.5\% to selection \#3 and so forth. This behaviour provides us with a skewed data set where the last selects receive only a small portion of the randomly generated samples. 
This problem is easily solved by our architecture as it uses an active learning algorithm, thus minimizing the effects of a skewed data set and allowing more efficient learning as described in algorithm \ref{alg:DNTtraining}.

\subsection{Distinguishing Tasks \& Learning Accurately}

\begin{figure}[ht!]\centering
    \begin{subfigure}{.5\columnwidth}
        \centering
        \includegraphics[width=1\columnwidth]{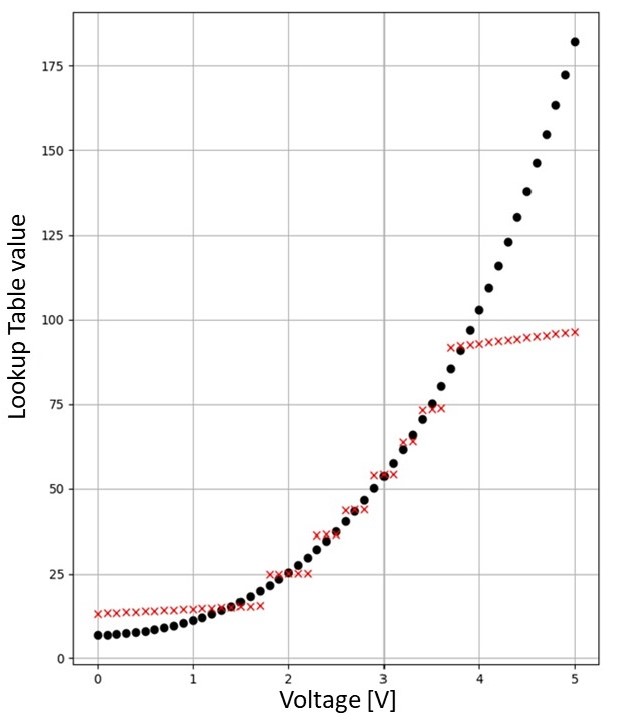}
        \caption{Select 2}
        \label{fig:EvaluationSel2_65iter}
    \end{subfigure}%
    \begin{subfigure}{.5\columnwidth}
        \centering
        \includegraphics[width=1\columnwidth]{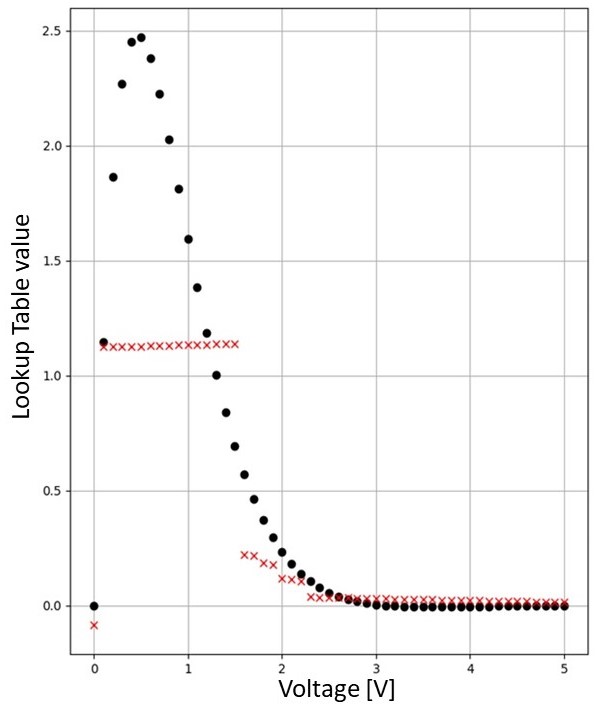}
        \caption{Select 3}
        \label{fig:EvaluationSel3_65iter}
    \end{subfigure}
    \caption{Prediction in red versus target in black: after only 65 iterations, black - target, red - prediction}
    \label{fig:EvaluationSelects65iter}
\end{figure}

\begin{figure}[ht!]\centering
    \begin{subfigure}{.5\columnwidth}
        \centering
        \includegraphics[width=1\columnwidth]{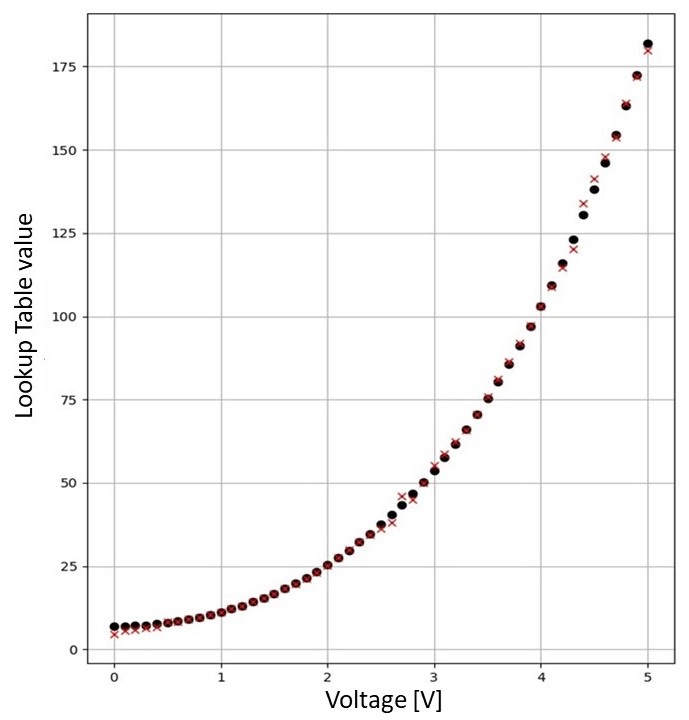}
        \caption{Select 2}
        \label{fig:EvaluationSel2_500iter}
    \end{subfigure}%
    \begin{subfigure}{.5\columnwidth}
        \centering
        \includegraphics[width=1\columnwidth]{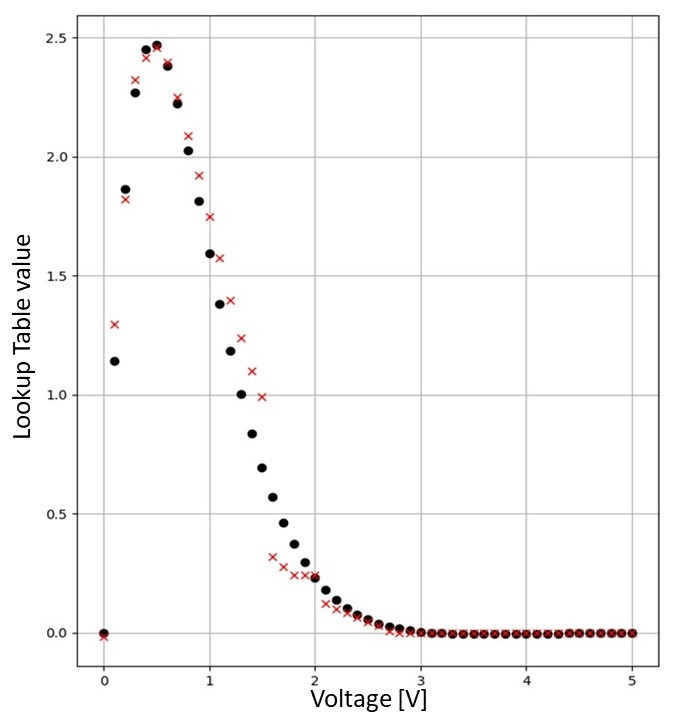}
        \caption{Select 3}
        \label{fig:EvaluationSel3_500iter}
    \end{subfigure}
    \caption{Prediction in red versus target in black: after only 500 iterations, black - target, red - prediction}
    \label{fig:EvaluationSelects500iter}
\end{figure}

We will describe which problems are relevant for the DNT architecture. Typical problem would be solved in the following manner: the depth of the regression tree would be corresponding to the task-separability of the problem. Thus, choosing the depth as $O(n)$ where $n$ is the number of separable tasks in the general hierarchy of the problem. In our case the depth of the regression tree is 8 while the number of possible tasks preformed by the chip is 18.
In addition, the size of each neural network connected to each leaf of the tree should be with the capacity of the sub-task we want to solve.

In Figures \ref{fig:EvaluationSelects65iter} \& \ref{fig:EvaluationSelects500iter} we test the architecture ability to distinguish two different tasks: select \#2 and select \#3.
Although the regression tree separates the task into smaller tasks, the NALU module is essential in order to learn complex and separated tasks.

\subsection{Time Efficient Sampling}

\begin{figure}[ht!]
    \includegraphics[width=\columnwidth]{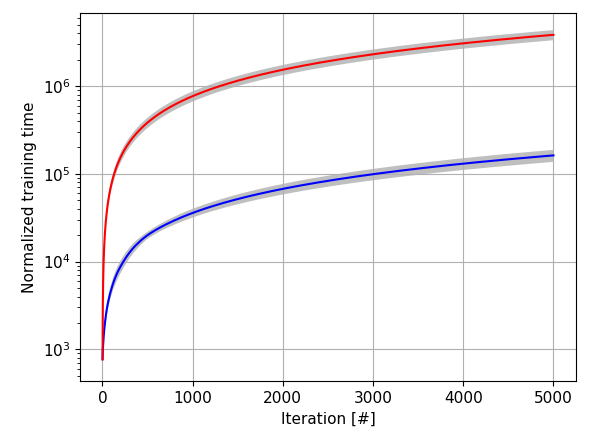}
    \caption{Normalized total cumulative training time with (blue line) and without (red line) the active learning algorithm, gray color represents one standard deviation. Normalization is done in comparison to batch training time, where batch size is 100. Training speed up of more than an order of magnitude was achieved. The results were averaged over 10 training procedures of the DNT architecture on the digital chip.}
    \label{fig:ComulativeVariance}
\end{figure}

\begin{figure}[!ht]
    \includegraphics[width=\columnwidth]{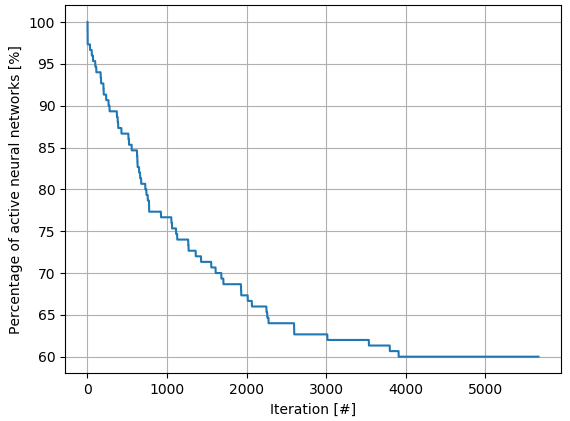}
    \caption{Iteration number versus percentage of active neural networks in a given training cycle. We can see that our active learning algorithm was able to stop the training of 40\% of the neural networks, thus reducing training time. $error = 10^{-3}$}
    \label{fig:DNTnetworkTrainingPercentage}
\end{figure}

As we can see in Figures \ref{fig:ComulativeVariance} \& \ref{fig:DNTnetworkTrainingPercentage}  as training continues more networks cease their training. A good example for efficient training appears at the first few training iterations; by stopping the training of a small percentage of the neural networks, we reduce the training time by a factor of 10 per iteration. A possible explanation for this behavior may be that in the first few iterations the architecture learns the "Enable" for input and output pins, thus learning which of the 18 inputs have the highest significance towards efficient learning and as result stop training the neural networks corresponding to those modes. In Figure \ref{fig:ComulativeVariance} we present the normalized training time in two cases. The first one is the cumulative sum of the training time of the architecture without the active learning algorithm and the second one is with it. As can be seen, the active learning algorithm speeds up the training time of the DNTs architecture in over one order of magnitude.

\section{Discussion \& Further Research} \label{Discussion & Further Research}
A step closer to full black-box cloning, DNT architecture practically relaxes all of the assumptions of previous black-box cloning scenarios, thus bridging the gap between theory and the practical world. Our architecture is a fully working module that could solve most simple black-boxes without memory with almost zero human interaction. The main advantage over most other methods of cloning comes from including the regression tree, therefore the cloning process could be explained, whereas most ML methods are inexplainable.
In the future we would like to extend the DNT architecture to handle black-boxes with internal memory. One may add layers that are suitable for learning time series data such as convolution, LSTM and others. Another approach would be to use the data stored within the black-box's memory as an additional input to the standard DNT. The scalability of the DNT will be able to handle this approach to some extent for simple tasks. Obviously, the problem will be more complex and the architecture of the attacking module will grow in size accordingly. The scalability problem may be reduced by using multiple node decision trees, or by hyperbolic neural networks as discussed by \cite{HyperbolicNeuralNetworks} which behave like tree structures that preserve hierarchy while losing explainability to some degree.

\newpage

\bibliography{mainbib}

\begin{thebibliography}{16}
\providecommand{\natexlab}[1]{#1}
\providecommand{\url}[1]{\texttt{#1}}
\expandafter\ifx\csname urlstyle\endcsname\relax
  \providecommand{\doi}[1]{doi: #1}\else
  \providecommand{\doi}{doi: \begingroup \urlstyle{rm}\Url}\fi

\bibitem[Breiman et~al.(1984)Breiman, Friedman, Olshen, and Stone]{Trees}
Breiman, L., Friedman, J., Olshen, R., and Stone, C. (eds.).
\newblock \emph{Classification and Regression Trees}.
\newblock Chapman and Hall/CRC, Wadsworth, Belmont, CA, 1984.

\bibitem[Ganea et~al.(2018)Ganea, Becigneul, and
  Hofmann]{HyperbolicNeuralNetworks}
Ganea, O., Becigneul, G., and Hofmann, T.
\newblock Hyperbolic neural networks.
\newblock \emph{Proc. of 32nd Conference on Neural Information Processing
  Systems}, 2018.

\bibitem[Jacovi et~al.(2019)Jacovi, Hadash, Kermany, Carmeli, Lavi, Kour, and
  Berant]{gradientBB}
Jacovi, A., Hadash, G., Kermany, E., Carmeli, B., Lavi, O., Kour, G., and
  Berant, J.
\newblock Neural network gradient-based learning of black-box function
  interfaces.
\newblock \emph{International Conference on Learning Representations}, 2019.

\bibitem[Li et~al.(2018)Li, Yi, and Zhang]{bbActiveLearning}
Li, P., Yi, J., and Zhang, L.
\newblock Query-efficient black-box attack by active learning.
\newblock \emph{arXiv preprint arXiv:1809.04913}, 2018.

\bibitem[Ma et~al.(2016)Ma, Destercke, and Wang]{ActiveLearningTrees}
Ma, L., Destercke, S., and Wang, Y.
\newblock Online active learning of decision trees with evidential data.
\newblock \emph{Pattern Recognition}, 2016.

\bibitem[Oh et~al.(2018)Oh, Augustin, Schiele, and Fritz]{ReverseBB}
Oh, S.~J., Augustin, M., Schiele, B., and Fritz, M.
\newblock Towards reverse-engineering black-box neural networks.
\newblock \emph{International Conference on Learning Representations}, 2018.

\bibitem[Orekondy et~al.(2019)Orekondy, Schiele, and Fritz]{Knockoff}
Orekondy, T., Schiele, B., and Fritz, M.
\newblock Knockoff nets: Stealing functionality of black-box models.
\newblock \emph{IEEE Conference on Computer Vision and Pattern Recognition},
  2019.

\bibitem[Papernot et~al.(2017)Papernot, McDaniel, Goodfellow, Jha, Celik, and
  Swami]{PracticalBB}
Papernot, N., McDaniel, P., Goodfellow, I., Jha, S., Celik, Z.~B., and Swami,
  A.
\newblock Practical black-box attacks against machine learning.
\newblock \emph{Asia CCS}, 2017.

\bibitem[Settles(2010)]{ActiveLearningSurvey}
Settles, B.
\newblock Active learning literature survey.
\newblock Technical report, University of Wisconsin-Madison, 2010.

\bibitem[Shokri et~al.(2017)Shokri, Stronati, Song, and Shmatikov]{Membership}
Shokri, R., Stronati, M., Song, C., and Shmatikov, V.
\newblock Membership inference attacks against machine learning models.
\newblock \emph{Security and Privacy}, 2017.

\bibitem[Sirat \& Nadal(1990)Sirat and Nadal]{NeuralTreeClass}
Sirat, J. and Nadal, J.
\newblock Neural trees: a new tool for classification.
\newblock \emph{Network: Computation in Neural Systems}, 1990.

\bibitem[Tanno et~al.(2019)Tanno, Arulkumaran, Alexander, Criminisi, and
  Nori]{AdaptiveNeuralTree}
Tanno, R., Arulkumaran, K., Alexander, D.~C., Criminisi, A., and Nori, A.
\newblock Adaptive neural trees.
\newblock \emph{Proceedings of the 36th International Conference on Machine
  Learning}, 2019.

\bibitem[Tjoa \& Guan(2019)Tjoa and Guan]{SurveyExplainableAI}
Tjoa, E. and Guan, C.
\newblock A survey on explainable artificial intelligence (xai): Towards
  medical xai.
\newblock \emph{arXiv preprint arXiv:1907.07374}, 2019.

\bibitem[Tramer et~al.(2016)Tramer, Zhang, Juels, Reiter, and
  Ristenpart]{SecurityPrivacy}
Tramer, F., Zhang, F., Juels, A., Reiter, M.~K., and Ristenpart, T.
\newblock Stealing machine learning models via prediction apis.
\newblock \emph{USENIX Security}, 2016.

\bibitem[Trask et~al.(2018)Trask, Hill, Reed, Rae, Dyer, and P.blunsom]{NALU}
Trask, A., Hill, F., Reed, S., Rae, J., Dyer, C., and P.blunsom.
\newblock Neural arithmetic logic units.
\newblock \emph{Proc. of 32nd Conference on Neural Information Processing
  Systems}, 2018.

\bibitem[Wang \& Gong(2018)Wang and Gong]{Hyperparamters}
Wang, B. and Gong, N.~Z.
\newblock Stealing hyperparameters in machine learning.
\newblock \emph{Security and Privacy}, 2018.

\end{thebibliography}
\bibliographystyle{icml2020}
\end{document}